\documentclass[letterpaper]{article} 
\usepackage[]{aaai23}
\usepackage{times}  
\usepackage{helvet}  
\usepackage{courier}  
\usepackage[hyphens]{url}  
\usepackage{graphicx} 
\usepackage{natbib}  
\usepackage{caption} 
\usepackage{algorithm}
\usepackage{algorithmic}
\usepackage{subfigure}
\usepackage{tikz}
\usepackage{amsmath}
\usepackage{amsthm}
\usepackage{booktabs}
\usepackage{amsfonts}
\usepackage{amssymb}
\usepackage{multicol}
\usepackage{multirow}
\usetikzlibrary{shapes.geometric}
\usepackage{newfloat}
\usepackage{listings}
\DeclareCaptionStyle{ruled}{labelfont=normalfont,labelsep=colon,strut=off} 
\floatstyle{ruled}
\newfloat{listing}{tb}{lst}{}
\floatname{listing}{Listing}

\title{Discovering New Intents Using Latent Variables}
\author {
    Yunhua Zhou,
    Peiju Liu,
    Yuxin Wang,
    Xipeng Qiu\thanks{\ \  Corresponding author.}\\
    }
\affiliations {
    School of Computer Science, Fudan University \\
    \{zhouyh20,xpqiu\}@fudan.edu.cn\\
    \{pjliu21, wangyuxin21\}@m.fudan.edu.cn \\
}

\usepackage{bibentry}

\begin{document}

\maketitle

\begin{abstract}
Discovering new intents is of great significance to establishing Bootstrapped Task-Oriented Dialogue System. Most existing methods either lack the ability to transfer prior knowledge in the known intent data or fall into the dilemma of forgetting prior knowledge in the follow-up. More importantly, these methods do not deeply explore the intrinsic structure of unlabeled data, so they can not seek out the characteristics that make an intent in general. In this paper, starting from the intuition that discovering intents could be beneficial to the identification of the known intents, we propose a probabilistic framework for discovering intents where intent assignments are treated as latent variables. We adopt Expectation Maximization framework for optimization. Specifically, In E-step, we conduct discovering intents and explore the intrinsic structure of unlabeled data by the posterior of intent assignments. In M-step, we alleviate the forgetting of prior knowledge transferred from known intents by optimizing the discrimination of labeled data. Extensive experiments conducted in three challenging real-world datasets demonstrate our method can achieve substantial improvements.
\end{abstract}

\section{Introduction}
\label{sec:intro}
Unknown intent detection~\cite{zhou-etal-2022-knn} in Bootstrapped Task-Oriented Dialogue System (BTODS) has gradually attracted more and more attention from researchers. However,~\textit{detecting} unknown intent is only the first step. For BTODS, ~\textit{discovering} new intents is not only the same basic but also more crucial and challenging. Because the preset intent set in BTODS is limited to cover all intents, BTODS should discover potential new intents actively during interacting with the users. 
Specifically, a large number of valuable unlabeled data will be generated within the interaction between users and the dialogue system. Considering the limited labeled corpus and time-consuming annotating, which also requires prior domain knowledge, the BTODS should adaptively identify known intents and discover unknown intents from those unlabeled data with the aid of limited labeled data.

Just as discovering new intents plays a crucial role in establishing BTODS, discovering new intents has raised a lot of research interest as unknown intent detection. Unsupervised cluster learning is one popular method to solve this problem.
To discover new intents from a large number of unlabeled data, many works~\cite{hakkani2013weakly,hakkani2015clustering,shi-etal-2018-auto,padmasundari2018intent} formalize this problem as an unsupervised clustering process. However, these methods mainly focus on how to construct pseudo-supervised signals to assist in guiding the clustering process and do not fully utilize the prior knowledge contained in the existing labeled data.

In a more general real scenario, we often have a small (but containing prior knowledge that can be used to guide the discovery of new intents) amount of labeled data in advance and a large amount of unlabeled data (e.g., in the dialogue scene mentioned above, it is generated in the interaction with the dialogue system), which contains both known intents and unknown intents to be discovered.
 Our purpose is to identify the known intents and discover the potential intents contained in the unlabeled corpus using labeled data.

Recently, ~\citet{lin2020discovering} propose that pairwise similarities can be used as pseudo supervision signals to guide the discovery of new intents. However, as in the analysis of ~\citet{zhang2021discovering}, this method can not achieve effective performance when there are more new intents to be discovered. Inspired by~\citet{caron2018deep}, \citet{zhang2021discovering} (DeepAligned) propose an effective method for discovering new intents. DeepAligned first finetunes the BERT~\cite{devlin2018bert} using the labeled data to transfer the prior knowledge and generalize the knowledge into the semantic features of unlabeled data. Further, to learn friendly representations for clustering, DeepAligned assigns a pseudo label to each unlabeled utterance and re-trains the model under the supervision of the softmax which is calculated by those pseudo labels.

\begin{figure}[htb]
    \centering
    \includegraphics[width=\linewidth]{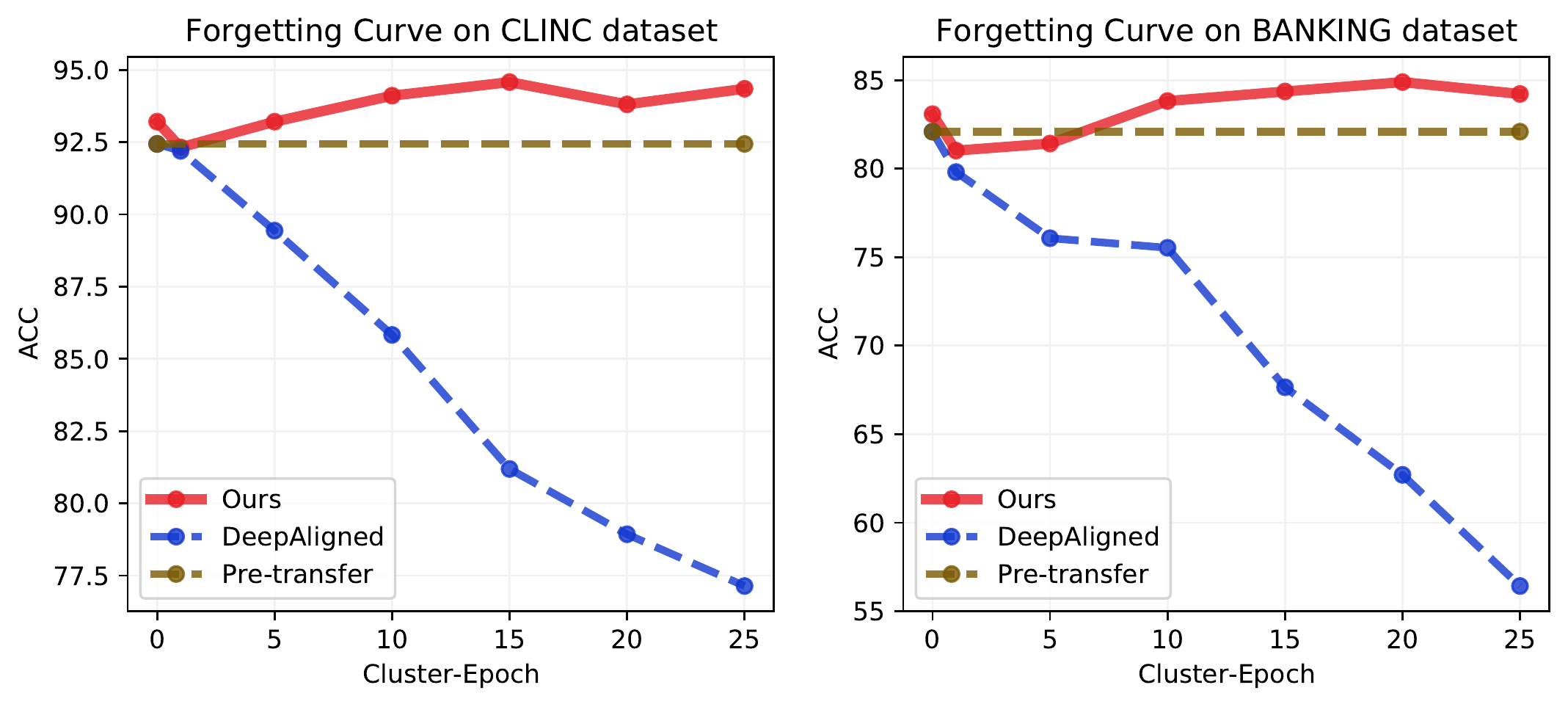}
    
    \caption{The forgetting curves of DeepAligned (Blue). 
    During discovering intents in DeepAligned, 
    the model constantly forgets the knowledge learned from labeled data. The brown line represents the baseline obtained by the model after transferring prior knowledge. In contrast, our method (Red) can alleviate forgetting well and See subsequent section for more discuss.}
    \label{fig:probknowledge}
\end{figure}

Nevertheless, DeepAligned may suffer from two critical problems. Firstly, when the model is re-trained with pseudo supervision signal, the model will forget the knowledge transferred in the transferring stage, the forgetting curves on different datasets as shown in Figure~\ref{fig:probknowledge}. During discovering intents in DeepAligned, we test the performance of the model on the validation set used in the transferring prior knowledge stage and show that with the advancement of clustering, the model constantly forgets the knowledge learned from labeled data. Furthermore, the model could be misled by inaccurate pseudo labels, particularly in large-sized intent space~\cite{wang2021self}.
More importantly, softmax loss formed by pseudo labels cannot explore the intrinsic structure of unlabeled data, so it can not provide accurate clustering supervised signals for discovering intents.

Different from the previous methods, we start from the essential intuition that the discovery of intents should not damage the identification of the known intents and the two processes should achieve a win-win situation. The knowledge contained in labeled data corpus can be used to guide the discovery of the new intents, and the information learned from the unlabeled corpus (in the process of discovering) could improve the identification of the known intents. 

Based on this intuition, with the help of optimizing identification of labeled data given the whole data corpus, we propose a principled probabilistic framework for intents discovery, where intent assignments as a latent variable. Expectation maximization provides a principal template for learning this typical latent variable model. Specifically, in the E-step, we use the current model to discover intents and calculate a specified posterior probability of intent assignments, which is to explore the intrinsic structure of data. In the M-step, maximize the probability of identification of labeled data (which is to mitigate catastrophic forgetting) and the posterior probability of intent assignments (which is to help learn friendly features for discovering new intents) simultaneously to optimize and update model parameters. Extensive experiments conducted in three benchmark datasets demonstrate our method can achieve substantial improvements over strong baselines. We summarize our contributions as follows:

\textbf{(Theory)} We introduce a principled probabilistic framework for discovering intents and provide a learning algorithm based on Expectation Maximization. To the best of our knowledge, this is the first complete theoretical framework in this field and we hope it can inspire follow-up research.

\textbf{(Methodology)} We provide an efficient implementation based on the proposed probabilistic framework. After transferring prior knowledge, we use a simple and effective method to alleviate the forgetting. Furthermore, we use the contrastive learning paradigm to explore the intrinsic structure of unlabeled data, which not only avoids misleading the model caused by relying on pseudo labels but also helps to better learn the features that are friendly to intent discovery.

\textbf{(Experiments and Analysis)} We conduct extensive experiments on a suite of real-world datasets and establish substantial improvements.

\section{Related Work}
\label{sec:related work}
Our work is mainly related to two lines of research: Unsupervised and Semi-supervised clustering.

\textbf{Unsupervised Clustering} Extracting meaningful information from unlabeled data has been studied for a long time. Traditional approaches like \textbf{K-means}~\cite{macqueen1967some} and Agglomerative Clustering \textbf{(AC)}~\cite{gowda1978agglomerative} are seminal but hardly perform well in high-dimensional space. Recent efforts are devoted to using the deep neural network to obtain good clustering representations. \citet{xie2016unsupervised} propose Deep Embedded Cluster \textbf{(DEC)} to learn and refine the features iteratively by optimizing a clustering objective based on an auxiliary distribution. Unlike DEC, \citet{yang2017towards} propose Deep Clustering Network \textbf{(DCN)} that performs nonlinear dimensionality reduction and k-means clustering jointly to learn friendly representation. \citet{chang2017deep} \textbf{(DAC)} apply unsupervised clustering to image clustering and proposes a binary-classification framework that uses adaptive learning for optimization. Then, \textbf{DeepCluster}~\cite{caron2018deep} proposes an end-to-end training method that performs cluster assignments and representation learning alternately. However, the key drawback of unsupervised methods is their incapability of taking advantage of prior knowledge to guide the clustering.

\textbf{Semi-supervised Clustering} With the aid of a few labeled data, semi-supervised clustering usually produces better results compared with unsupervised counterparts. \textbf{PCK-Means}~\cite{basu2004active} proposes that the clustering can be supervised by pairwise constraints between samples in the dataset. \textbf{KCL}~\cite{hsu2017learning} transfers knowledge in the form of pairwise similarity predictions firstly and learns a clustering network to transfer learning. Along this line, \textbf{MCL}~\cite{hsu2019multi} further formulates multi-classification as meta classification that predicts pairwise similarity and generalizes the framework to various settings. \textbf{DTC}~\cite{han2019learning} extends the DEC algorithm and proposes a mechanism to estimate the number of new images categories using labeled data. When it comes to the field of text clustering, \textbf{CDAC+}~\cite{lin2020discovering} combines the pairwise constraints and target distribution to discover new intents while \textbf{DeepAligned}~\cite{zhang2021discovering} introduces an alignment strategy to improve the clustering consistency. Very recently, \textbf{SCL}~\cite{shen2021semi} incorporates a strong backbone MPNet in the Siamese Network structure with contrastive loss (or rely on a large amount of additional external data~\cite{zhang-etal-2022-new}) to learn the better sentence representations. Although these methods take known intents into account, they may suffer from knowledge forgetting during the training process. More importantly, these methods are insufficient in the probe into the intrinsic structure of unlabeled data, making it hard to distinguish the characteristics that form an intent.

\section{Approach}
\subsection{Problem Definition}
Given as input an labeled dataset $D^l=\{x_i^l, i=1,\ldots,N\}$ where intents $Y^l=\{y_i^l, i=1,\ldots,N\}$ are known and an unlabeled dataset $D^u=\{x_i^u, i=1,\ldots,M\}$. Our goal is to produce intent assignments as output by clustering (or partition) the whole dataset $D$, which denotes $D={D^l}\cup{D^u}$. Directly optimizing the goal is intractable as the lack of knowledge about new intents and the intrinsic structure of unlabeled data. As analyzed in introduction, discovering intents should not damage but be beneficial for the identification of known intents, which can be formulated to optimize $p(Y^l|D^l, D;\theta)$.

Denote our latent variable representing intent assignments obtained by clustering on $D$ by $Z$ and let $\mathcal{Z}_D$ be a possible value of $Z$. Using Bayes rule, $p(Y^l|D^l, D;\theta)$ can be calculated as: 
\begin{equation}
    \begin{aligned}
        \label{eq:origin}
        p(Y^l|D^l,D) = \sum_{\mathcal{Z_D} \in Z}p(Y^l|\mathcal{Z}_D, D^l)p(\mathcal{Z}_D|D^l).
    \end{aligned}
\end{equation}

Exactly optimizing Eq.~\eqref{eq:origin} is intractable as it is combinatorial in nature. Consider a specific value $\mathcal{Z}$ (omitting subscript D for clarity), the log-likelihood can be simplified as:
\begin{align}
    \label{eq:obj}
    \mathcal{L}_{obj} = \log p(Y^l|\mathcal{Z}, D^l; \theta) + \log p(\mathcal{Z}|D^l; \theta).
\end{align}
Our goal is get better $\mathcal{Z}$ (i.e.intent discovery) by optimizing $\mathcal{L}_{obj}$, and a better $\mathcal{Z}$ can also help optimize $\mathcal{L}_{obj}$.

\subsection{Intent Representation and Transfer Knowledge}
Before optimizing $\mathcal{L}_{obj}$, we want to transfer knowledge from labeled corpus to initialize the model. Transferring knowledge has been widely studied and types of transferred knowledge have been proposed for a variety of circumstances. Considering the excellent generalization of the pre-trained model, we fine-tune BERT~\cite{devlin2018bert} with labeled corpus under the supervision of cross entropy.
Given the i-th labeled utterance, we first get its contextual embeddings $[[CLS],T_1,T_2,...,T_N]$ by utilizing BERT and then perform mean-pooling to get sentence semantic representation $Z_i= \text{Mean-Pooling}([[\text{CLS}], T_1,... T_N])$, where $Z_{i}\in \mathcal{R}^{H}$, N is the sequence length and H is the hidden dimension. 
The objective of fine-tune $\mathcal{L}_{ce}$ as:
\begin{align}
    \label{eq:ce}
    \mathcal{L}_{\text{ce}} =  -\frac{1}{N}\sum_{i=1}^{N}\log\frac{\exp(\phi(z_i)^{y_i})}{\sum_{j=1}^{K^{l}}\exp(\phi(z_i)^j)},
\end{align}
where \begin{math}\phi(\cdot)\end{math} denotes linear classifier and \begin{math}\phi(z_i)^j\end{math} denotes the logits of the j-th class.

\subsection{EM Framework for Optimization}
\label{sec:EM framework}
\textbf{Intent Assignments $\mathcal{Z}$} Specific intent assignments $\mathcal{Z}$ involves two components: how to determine K representing how many intents in dataset $D$ and how to assign the utterance in the dataset to corresponding intent.
Many methods ~\cite{han2019learning,shen2021semi} have been proposed to estimate K. Considering the tradeoff between the efficiency and effect, we follow~\citet{zhang2021discovering} (see subsequent analysis for improvements by us). We first set a rough value $\mathcal{K}$ (e.g., the multiple of the ground truth number) for K and extract the semantic feature of the utterance using above fine-tuned BERT. Furtherly, we group the dataset $D$ as K semantic clusters using k-means and drop clusters whose size is less than a certain threshold. The K is calculated as:
\begin{align}
    K = \sum_{i=1}^{\mathcal{K}}\mathbb{I}(|C_i|>=\theta),
\end{align}
where $|C_i|$ is the size of i-th produced cluster, $\mathbb{I}$ is an indicator function and $\theta$ is the threshold which is set to the same value as suggested in~\citet{zhang2021discovering}.

After estimating how many intents are contained in the dataset, we perform k-means to assign cluster assignments as (pseudo) intent to each utterance. Next, We discuss in detail how to further optimize Eq.~\eqref{eq:obj} with Expectation-Maximization (EM) algorithm framework. 
\begin{figure*}[!ht]
    \centering
    \includegraphics[width=0.95\linewidth]{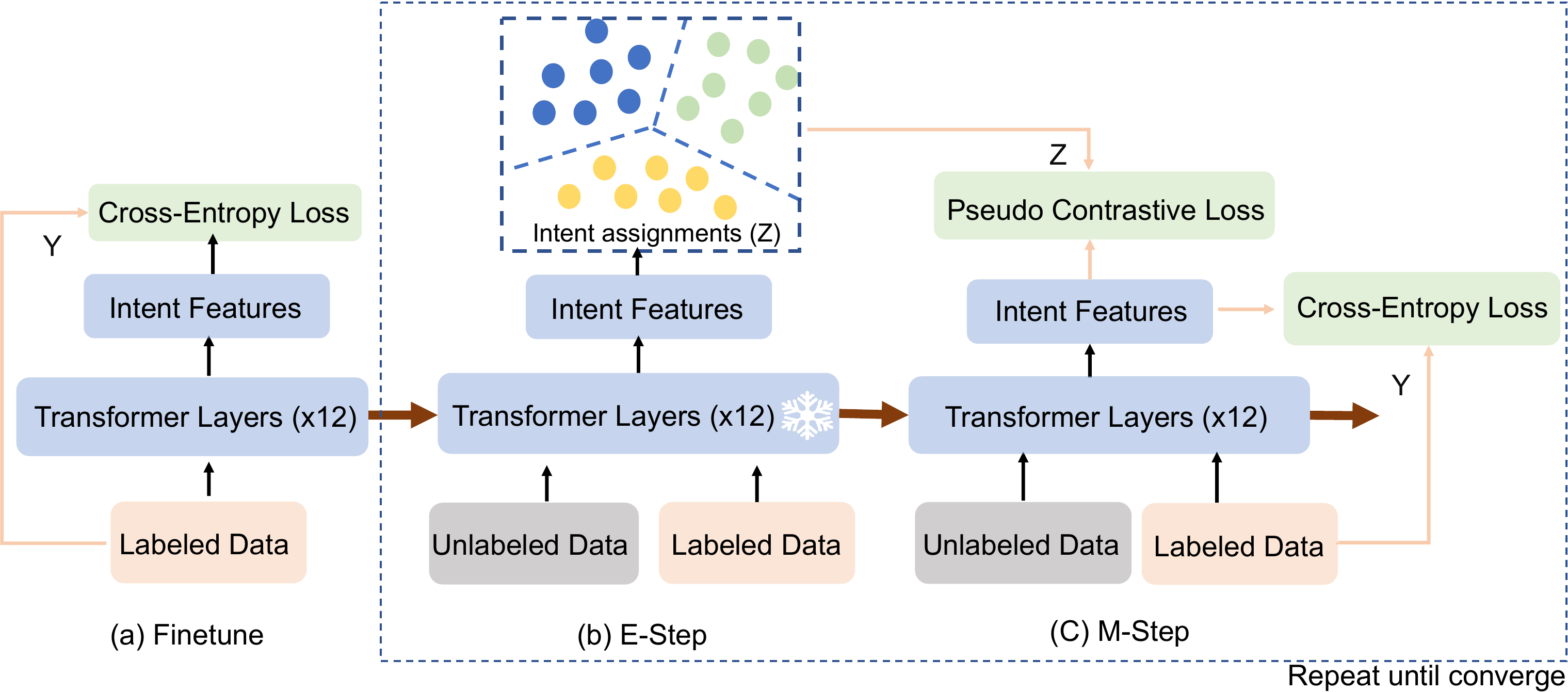}
    \caption{The model architecture of our implementation based on proprosed probabilistic framework. (a) Firstly, we transfer knowledge by fine-tuning BERT with labeled data. (b) Then, we perform intent assignments on full data (labeled and unlabeled data) and reflect the intrinsic structure of data in E-step. (c)And alleviate the forgetting of prior knowledge and update model parameters in M-step. The snow mark represents this step only needs forward without calculating the gradient.}
    \label{fig:architecture}
\end{figure*}

\textbf{E-Step} We have assigned a specific intent assignment $\mathcal{Z}$ to latent variable $Z$ based on prior knowledge. We expect that the intent assignments $\mathcal{Z}$ should reflect what characteristics make a good intent in general rather than specific intents. Therefore, the standard cross entropy loss formed by specific pseudo labels adopted by~\citet{caron2018deep,zhang2021discovering} can not achieve this purpose, and even the model may be confused by the false pseudo labels according to~\citet{wang2021self}. To better reflect the intrinsic structure of dataset $D$ and learn friendly features for intent assignments, we hope that intent assignments $\mathcal{Z}$ can make utterances with the same intent close enough and pull utterances with different intents far away in the semantic feature space. Inspired by contrastive learning paradigm, we estimate the posterior $p(\mathcal{Z}|D^l;\theta)$:
\begin{align}
    \label{eq:post}
    p(\mathcal{Z}|D^l; \theta) =& \prod_{C_k\in \mathcal{Z}}p(C_k|D^l; \theta) \\
                             =& \prod_{C_k\in \mathcal{Z}}\prod_{x\in C_k}p(x\in C_k|D^l; \theta) \\
                            \propto & \prod_{C_k\in \mathcal{Z}}\prod_{x\in C_k}\frac{\sum_{x^+\in C_k}exp(x\cdot x^+)}{\sum_{x^p\in D\backslash\{x\}}exp(x\cdot x^p)},
\end{align}
where $C_k$ is a cluster produced by $\mathcal{Z}$, and $x\cdot x^+$ is calculated by consine between features. To optimize Eq.~\eqref{eq:obj}, we also need to compute $p(Y^l|\mathcal{Z}, D^l; \theta)$. Exactly computing 
is difficult as the label space in $\mathcal{Z}$ does not match that of $Y^l$. Consider the disaster forgetting as in Deepaligned mentioned above, we approximate $p(Y^l|\mathcal{Z}, D^l; \theta)$:
\begin{align}
    \label{eq:prior}
    p(Y^l|\mathcal{Z}, D^l; \theta) \propto& \quad p(Y^l|D^l; \theta) \\
    \propto& \prod_{x\in D^l}\frac{exp(\phi(x)^{y})}{\sum_{j=1}^{K^l}exp(\phi(x)^j)},
\end{align}
where \begin{math}\phi(\cdot)\end{math} denotes same linear classifier as Eq.~\eqref{eq:ce}, $y$ denotes the intent of $x$, $K^l$ denotes the total number of known intents and $D^l$ denotes labeled data in $D$. 

Our goal is to tailor the labeled data into model training. On the one hand, the model will not lose the knowledge transferred from labeled data, on the other hand, the model can constantly explore the intrinsic structure of the dataset by utilizing it. 

\textbf{M-Step} In the M-step, we update the $\theta$ in Eq.~\eqref{eq:obj}. In addition to bring Eq.~\eqref{eq:post} and Eq.~\eqref{eq:prior} into Eq.~\eqref{eq:obj}, we introduce two hyper-parameters to help optimize objectives. The overall loss $\mathcal{L}$ can be formulated as follows:
\begin{align}
    \label{eq:loss}
    \mathcal{L} &= \lambda \cdot \sum_{C_k\in \mathcal{Z}}\sum_{x\in C_k} \log\frac{\sum_{x^+\in C_k}exp(\frac{x\cdot x^+}{\tau})}{\sum_{x^p\in D\backslash\{x\}}exp(\frac{x\cdot x^p}{\tau})} \\
                &+ (1-\lambda) \cdot \sum_{x\in D^l} \log\frac{exp(\phi(x)^y)}{\sum_{j=1}^{K^l}exp(\phi(x)^j)},
\end{align}
where $\lambda$ is to balance the proportion of two log-likelihoods during training, $\tau$ is a hyper-parameter for temperature scaling which often appears in contrastive learning. 

\begin{algorithm}[!ht]
\caption{EM algorithm for optimization}
\label{alg:algorithm}
\textbf{Input}: $D^l=\{x_i^l, i=1,\ldots,N\}$, $Y^l=\{y_i^l, i=1,\ldots,N\}$, $D^u=\{x_i^u, i=1,\ldots,M\}$.\\
\textbf{Parameter}: Model parameters $\theta$.

\begin{algorithmic}[1] 
\STATE Intialize $\theta$ by transferring knowledge.
\WHILE{not converged}
\STATE Perform intent assignment $\mathcal{Z}$ using K-means; $\backslash$$\backslash$ E-Step
\STATE Compute $P(Y^l|\mathcal{Z}, D^l; \theta)$ and $P(\mathcal{Z}|D^l;\theta)$ using current parameters $\theta$; $\backslash$$\backslash$ E-Step
\STATE Update model parameters $\theta$ to maximize the log-likelihood in Eq.~\eqref{eq:loss}. $\backslash$$\backslash$ M-Step
\ENDWHILE
\STATE \textbf{return} $\theta$
\end{algorithmic}
\end{algorithm}

We summarize the whole training process of EM framework in Algorithm 1 and the model architecture of our approach as shown in Figure~\ref{fig:architecture}. It is worth noting that our method actually proposes a framework where probability estimation can flexibly adopt different ways for a variety of circumstances.

\begin{table*}[!ht]
    \centering
    \begin{tabular}{l|c|c|c|c|c|c}
    \toprule
    Dataset & Classes & \#Training & \#Validation & \#Test & Vocabulary Size & Length (Avg) \\
    \midrule
     CLINC & 150 & 18000 & 2250 & 2250 & 7283 & 8.32  \\
     BANKING & 77 & 9003 & 1000 & 3080 & 5028 & 11.91 \\
     StackOverflow & 20 & 12000 & 2000 & 6000 & 17182 & 9.18 \\
    \bottomrule
    \end{tabular}
    \caption{Statistics of datasets. \# denotes the total number of utterances.}
    \label{tab:stastic}
\end{table*}

\section{Experiments}
\subsection{Datasets}
We conduct experiments on three challenging datasets to verify the effectiveness of our proposed method. The detailed statistics are shown in Table~\ref{tab:stastic}.

\textbf{CLINC}~\cite{larson2019evaluation} is a dataset designed for Out-of-domain intent detection, which contains 150 intents from 10 domains and 22500 utterances.

\textbf{BANKING}~\cite{casanueva2020efficient} is a dataset covering 77 intents and containing 13083 utterances.

\textbf{StackOverflow} is a dataset published in Kaggle.com, which has 20 intents and 20000 utterances. We adopt the dataset processed by ~\citet{xu2015short}.

\subsection{Baseline and Evaluation Metrics}
We follow~\citet{lin2020discovering,zhang2021discovering} and divide the baselines to be compared into two categories: Unsupervised (Unsup.) and Semi-supervised (Semi-sup.). All methods are introduced in Related Work. For a fairness, we uniformly use BERT as the backbone network when compared with the above methods. We also note that SCL~\cite{shen2021semi} uses stronger backbone network to obtain semantically meaningful sentence representations, and we also use the same backbone network in comparison with these methods.

To evaluate clustering results, we follow existing methods~\cite{lin2020discovering,zhang2021discovering} and adopt three widely used metrics: Normalized Mutual Information (NMI), Adjusted Rand Index (ARI), and clustering accuracy (ACC). It should be noted that when calculating ACC, the Hungarian algorithm is adopted to find the optimal alignment between the pseudo labels and the ground-truth labels as following~\citet{zhang2021discovering}.
\begin{table*}[!ht]
    \small
    \centering
    \begin{tabular}{l l | c c c | c c c | c c c}
    \toprule
     & \multirow{3}{*}{Methods} & \multicolumn{3}{c}{CLINC} & \multicolumn{3}{c}{BANKING} & \multicolumn{3}{c}{StackOverflow} \\
    \cmidrule{3-5} \cmidrule{6-8} \cmidrule{9-11}
     &  & NMI & ARI & ACC & NMI & ARI & ACC & NMI & ARI & ACC \\
    \midrule
    \multirow{7}*{Unsup.} & K-means & 70.89 & 26.86 & 45.06 & 54.57 & 12.18 & 29.55 & 8.24 & 1.46 & 13.55 \\
    ~ & AC & 73.07 & 27.70 & 44.03 & 57.07 & 13.31 & 31.58 & 10.62 & 2.12 & 14.66\\
    ~ & SAE-KM & 73.13 & 29.95 & 46.75 & 63.79 & 22.85 & 38.92 & 32.62 & 17.07 & 34.44\\
    ~ & DEC & 74.83 & 27.46 & 46.89 & 67.78 & 27.21 & 41.29 & 10.88 & 3.76 & 13.09 \\
    ~ & DCN & 75.66 & 31.15 & 49.29 & 67.54 & 26.81 & 41.99 & 31.09 & 15.45 & 34.56\\
    ~ & DAC & 78.40 & 40.49 & 55.94 & 47.35 & 14.24 & 27.41 & 14.71 & 2.76 & 16.30 \\
    ~ & DeepCluster & 65.58 & 19.11 & 35.70 & 41.77 & 8.95 & 20.69 & - & - & - \\
    \midrule
    \midrule
    \multirow{6}*{Semi-sup.} & PCKMeans & 68.70 & 35.40 & 54.61 & 48.22 & 16.24 & 32.66 & 17.26 & 5.35 & 24.16\\
    ~ & KCL(BERT) & 86.82 & 58.79 & 68.86 & 75.21 & 46.72 & 60.15 & 8.84 & 7.81 & 13.94\\
    ~ & MCL(BERT) & 87.72 & 59.92 & 69.66 & 75.68 & 47.43 & 61.14 & - & - & -\\
    ~ & CDAC+ & 86.65 & 54.33 & 69.89 & 72.25 & 40.97 & 53.83 & 69.84 & 52.59 & 73.48\\
    ~ & DTC(BERT) & 90.54 & 65.02 & 74.15 & 76.55 & 44.70 & 56.51 & - & - & - \\
    ~ & DeepAligned & 93.89 & 79.75 & 86.49 & 79.56 & 53.64 & 64.90 & 76.47 & 62.52 & 80.26 \\
    \midrule
    ~ & \textit{Ours} & \textbf{95.13}\textsubscript{0.46}& \textbf{82.65}\textsubscript{1.77}& \textbf{88.35}\textsubscript{1.22}& \textbf{83.40}\textsubscript{1.44}& \textbf{61.19}\textsubscript{3.15}& \textbf{72.59}\textsubscript{1.77}&
    \textbf{77.29}\textsubscript{0.80}& \textbf{63.93}\textsubscript{2.71}& \textbf{80.90}\textsubscript{1.28}\\
    \bottomrule
    \end{tabular}
    \caption{The main results on three datasets. The baselines on CLINC and BANKING are retrieved from~\protect\citet{zhang2021discovering}. The baselines on StackOverflow are retrieved from~\protect\citet{lin2020discovering}. We get the baseline of DeepAligned on StackOverflow by running its release code.
    All reported results are percentages and mean by conducting with different seeds (The subscripts are the corresponding standard deviations).}
    \label{tab:bert results}
\end{table*}

\subsection{Experimental Settings}
For each dataset, we randomly select 75\% of all intents as known and regard the remaining as unknown. Furthermore, we randomly choose 10\% of the known intents data as labeled data. We set the number of intents as ground-truth. Our experimental settings is the same as ~\citet{lin2020discovering,zhang2021discovering} for fair comparison. We take different random seeds to run three rounds on the test set and report the averaged results.

Our main experiments use pre-trained BERT (bert-uncased, with 12-layer transformer), which is implemented in PyTorch, as the network backbone. We also replace the backbones of the compared baselines with the same BERT as ours.
We adopt most of the suggested hyper-parameters for learning. We try learning rate in $\{1e-5, 5e-5\}$ and $\lambda$ in $\{0.5, 0.6\}$. The training batch size is 256, and the temperature scale $\tau$ is 0.1. All experiments were conducted in the Nvidia Ge-Force RTX-3090 Graphical Card with 24G graphical memory.
Only when comparing with SCL~\cite{shen2021semi}, which definitely point out that they use pre-trained MPNet~\cite{reimers2019sentence} as the backbone network, will we adopt the same backbone network for a fair comparison.

Moreover, considering the efficiency of the training process and the capacity of GPU, we only fine-tune the last transformer layer parameters during transferring knowledge and freeze all but the latter 6 transformer layers parameters during performing EM algorithm.
\begin{table}[!th]
    \small
    \centering
    \setlength\tabcolsep{4pt}
    \begin{tabular}{l | c c c | c c c}
    \toprule
     \multirow{3}{*}{Methods} & \multicolumn{3}{c}{CLINC} & \multicolumn{3}{c}{BANKING} \\
    \cmidrule{2-4} \cmidrule{5-7} 
    ~ & NMI & ARI & ACC & NMI & ARI & ACC \\
    \midrule
    SMPNET & 93.39 & 74.28 & 83.24 & 82.22 & 58.82 & 71.82 \\
    SCL & 94.75 & 81.64 & 86.91 & 85.04 & 65.43 & 76.55 \\
    SCL(EP) & 95.25 & 83.44 & 88.68 & 84.77 & 64.44 & 75.18 \\ 
    SCL(IP) & 94.95 & 82.32 & 88.28 & 84.82 & 64.51 & 74.81 \\
    SCL(AA) & 95.11 & 83.09 & 88.49 & 85.02 & 64.91 & 75.66 \\
    SCL(AC) & 94.04 & 78.99 & 84.58 & 83.52 & 62.18 & 73.09 \\
    \midrule
    \textit{Ours} & \textbf{95.69} & \textbf{84.81} & \textbf{89.57} & \textbf{85.97} & \textbf{67.54} & \textbf{76.82}\\
    \bottomrule
    \end{tabular}
    \caption{The results compared with SCL and variants. IP, EP, AA, and AC represent four pseudo label training strategies:inclusive pairing, exclusive pairing, Alignment-A, and Alignment-C respectively. The baselines are retrieved from~\protect\citet{shen2021semi}.}
    \label{tab:mp-net results}
\end{table}

\section{Results and Discussion}
\subsection{Main results}
We present the main results in table~\ref{tab:bert results}, where the best results are highlighted in bold. It is clear from the results that our method achieves substantial improvements in all metrics and all datasets, especially in the BANKING dataset, where the number of samples in each class is imbalanced. These results illustrate the effectiveness and generalization of our method. 
At the same time, we note most semi-supervised methods are better than unsupervised as a whole, which further verifies the importance of labeled data.
From this perspective, we can explain why our method can be better than DeepAligned as it will constantly forget the knowledge existing in labeled data as shown in Introduction, and our method tailors the labeled data into model training to guide clustering so that our method can achieve better results.

To make a fair comparison with SCL~\cite{shen2021semi}, we also replace the backbone network in our method with the same MPNet as SCL, keeping other parts of our method unchanged. We present the results of our comparison with SCL and various variants (See~\citet{shen2021semi} for the calculation of specific strategies) on CLINC and BANKING in Table~\ref{tab:mp-net results}, where the best results are also highlighted in bold.
\begin{figure}[htb]
    \centering
    \includegraphics[width=\linewidth]{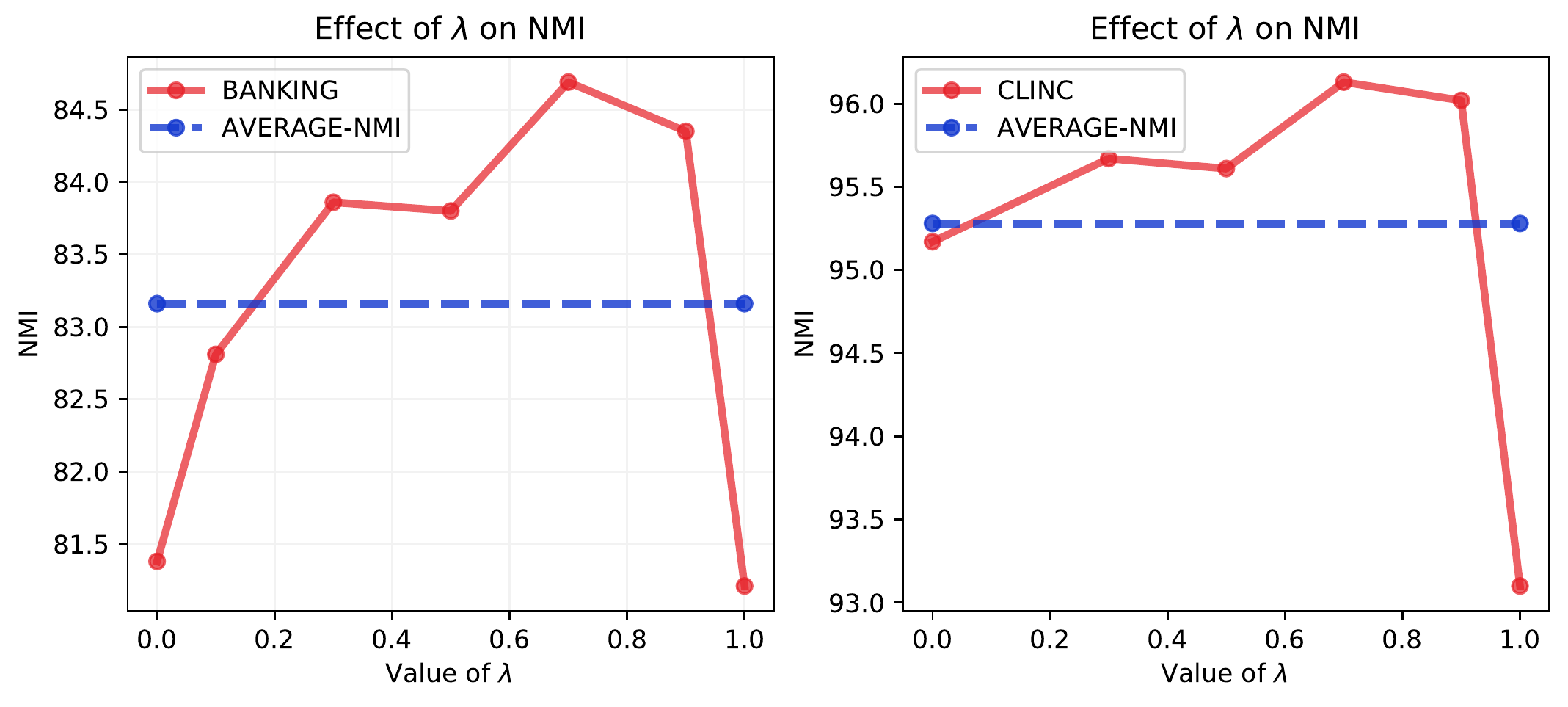}
    \includegraphics[width=\linewidth]{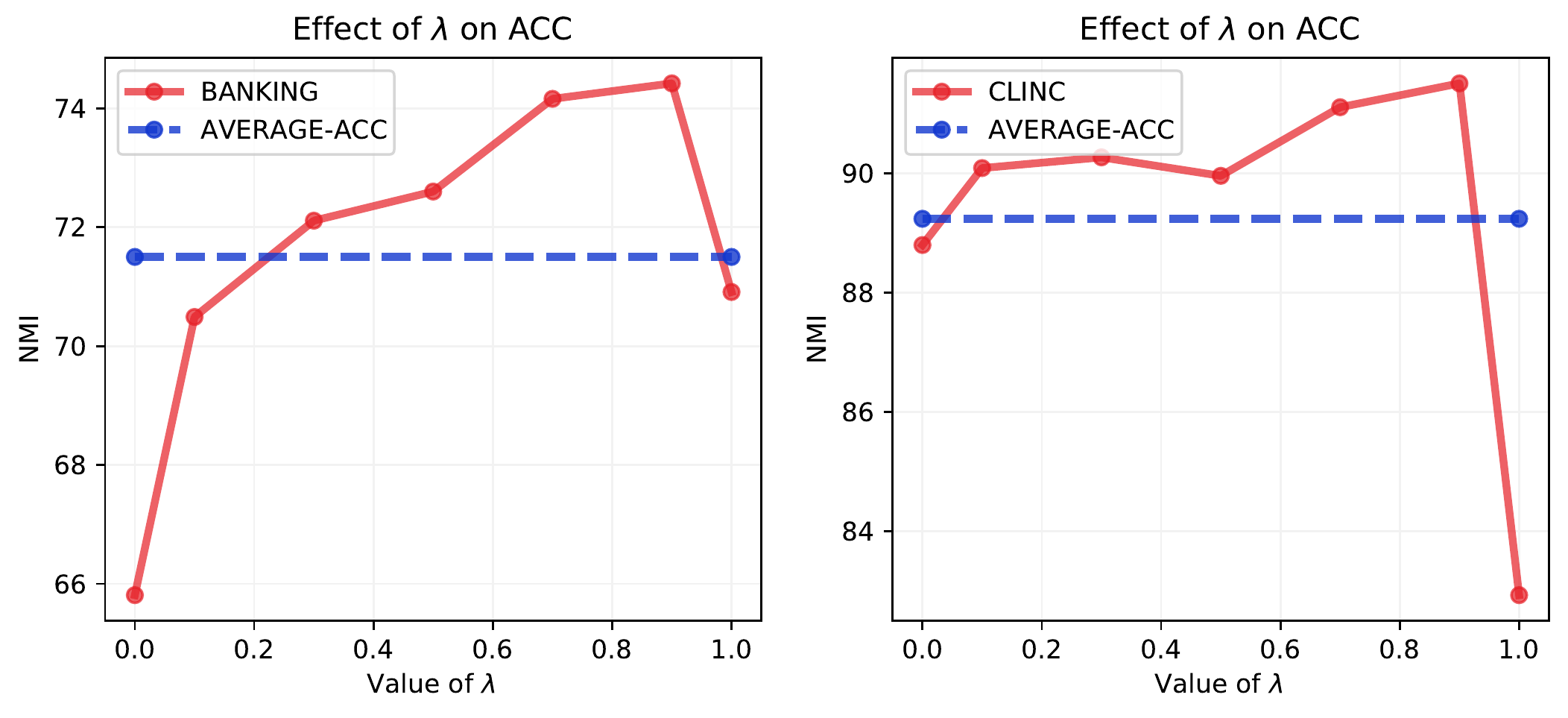}
    \caption{The effects of $\lambda$ on datasets (Right: CLINC, Left: BANKING). Only utilizing labeled data or only exploring the intrinsic structure will not achieve good results.}
    \label{fig:lambda}
\end{figure}

\subsection{Effect of Exploration and Utilization}
\label{sec:lambel}
In objective function Eq.~\eqref{eq:loss}, we use $\lambda$ to reconcile the effects of the two log-likelihoods. Intuitively, the first term is used to explore the intrinsic structure of unlabeled data, and the second term is used to strengthen the knowledge transferred from labeled data to utilize. We vary the value of $\lambda$ and conduct experiments on CLINC and BANKING to explore the effect of $\lambda$, which also reflects the inference of exploration and utilization.
As shown in Figure~\ref{fig:lambda}, only utilizing labeled data ($\lambda$ = 0.0) or only exploring($\lambda$ = 1.0) the intrinsic structure will not achieve good results (below average). Interestingly, on all metrics and datasets, the effect of $\lambda$ shows a similar trend (increase first and then decrease), which indicates that we can adjust the value of $\lambda$ to give full play to the role of both so that the model can make better use of known knowledge to discover intents accurately. This result shows that if the model wants to achieve good results, exploration and utilization are indispensable.
\begin{table}[ht]
    \small
    \centering
    \setlength\tabcolsep{2pt}
    \begin{tabular}{l | c c c | c c c}
    \toprule
     \multirow{3}{*}{Methods} & \multicolumn{3}{c}{CLINC ($\hat{K}$ = 150)} & \multicolumn{3}{c}{BANKING ($\hat{K}$ = 77)} \\
    \cmidrule{2-4} \cmidrule{5-7} 
    ~ & K & Error $\downarrow$ & ACC $\uparrow$ & K & Error $\downarrow$ & ACC $\uparrow$ \\
    \midrule
    MCL(BERT) & 112 & 25.33 & 69.2 & 58 & 24.68 & 60.8 \\
    DTC(BERT) & 195 & 30.00 & 66.65 & 110 & 42.86 & 54.94 \\
    DeepAligned & 129 & 14.00 & 77.18 & 67 & 12.99 & 62.49 \\ 
    \midrule
    \textit{Ours} & \textbf{130} & \textbf{13.3} & \textbf{80.8} & \textbf{73} & \textbf{5.48} & \textbf{69.68}\\
    \bottomrule
    \end{tabular}
    \caption{The results of predicting K. The $\hat{K}$ denotes the ground truth number of K. The closer $\hat{K}$ and K is, the more accurate the prediction is.  The compared results are retrieved from~\protect\citet{zhang2021discovering}.}
    \label{tab:results of 2k}
\end{table}
\subsection{Estimate the Number of Intents (K)}
\label{sec:estimate intents}
A key point of intent discovery is whether the model can accurately predict the number of intents. DeepAligned proposes a simple yet effective estimation method. However, due to the alignment operation in the iterative process of clustering (see ~\citet{zhang2021discovering} for details), DeepAligned needs to determine K in advance and only limited labeled data is used, while a large number of unlabeled data are ignored. Unlikely, our method does not directly rely on pseudo labels so that we can continue to refine K during subsequent clustering. We use the same settings as ~\citet{zhang2021discovering} and firstly assign the number of intents (i.e., $\mathcal{K}$ in intent assignments) as two times the ground truth number to investigate the ability to estimate K. In the process of executing EM algorithm, we refine K per 10 epochs using the method as suggested in above Section. 
We get the final performance of the model and the results are shown in Table~\ref{tab:results of 2k} show that our method can predict the number of intents more accurately and achieve better results at the same time.
\begin{figure}[htb]
    \centering
    \includegraphics[width=\linewidth]{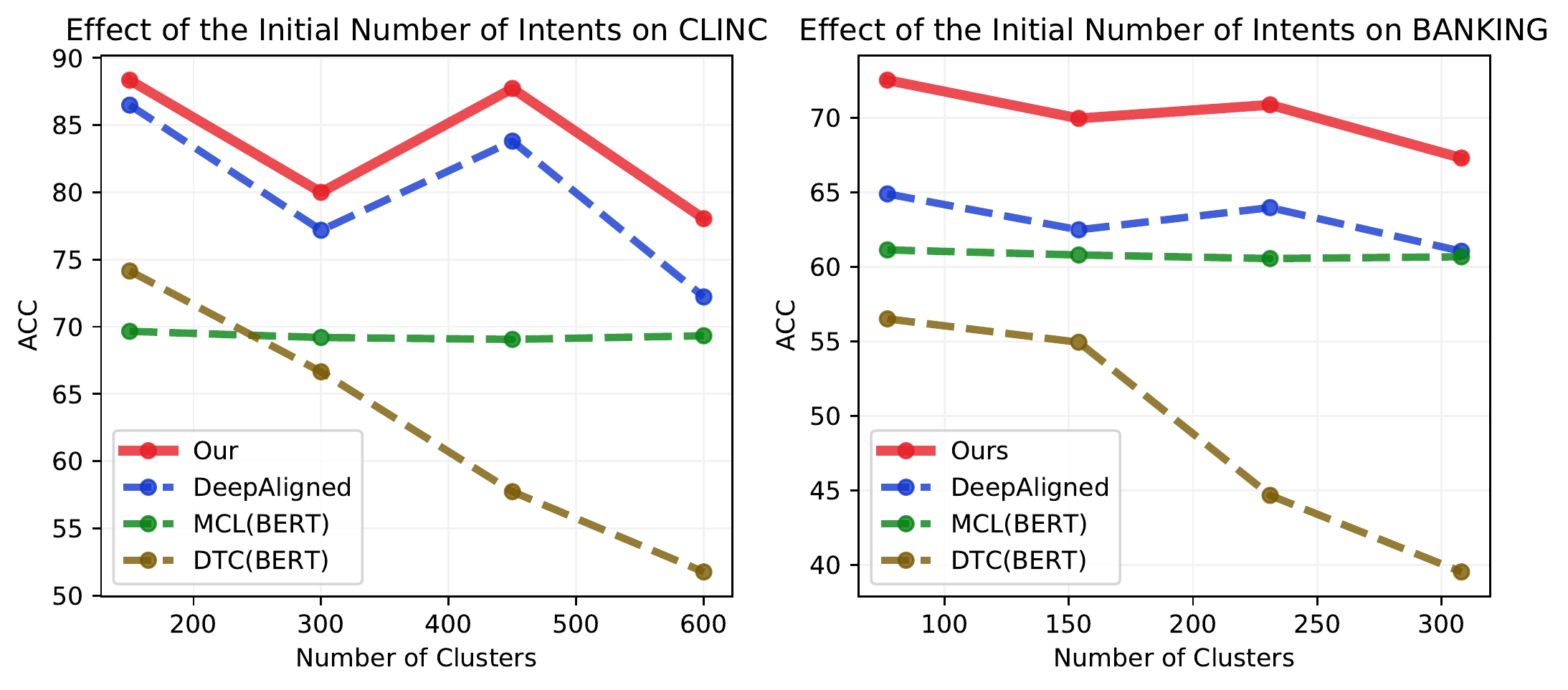}
    \caption{The effect of the Initial Number of Intents on datasets(Left: CLINC, Right: BANKING). The compared results are retrieved from~\protect\citet{zhang2021discovering}.}
    \label{fig:clusters-k}
\end{figure}

\subsection{Effect of the Initial Number of Intents}
\label{sec:number of intents}
Because we do not know the actual number of intents, we usually need to assign an initial number of intents (i.e., $\mathcal{K}$) in advance as we do earlier. This also requires us to investigate the sensitivity of the model to the initial K. We investigate the performance of our method in the datasets by varying initial values (leaving others unchanged). As shown Figure~\ref{fig:clusters-k}, compared with others, our method can better adapt to different initial values. Combined with the experiments in previous section,
we suppose the main reason why our method can achieve better performance is that our method can refine K more accurately than other methods by making the most of the knowledge of the whole data. 

\begin{figure}[htb]
    \centering
    \includegraphics[width=\linewidth]{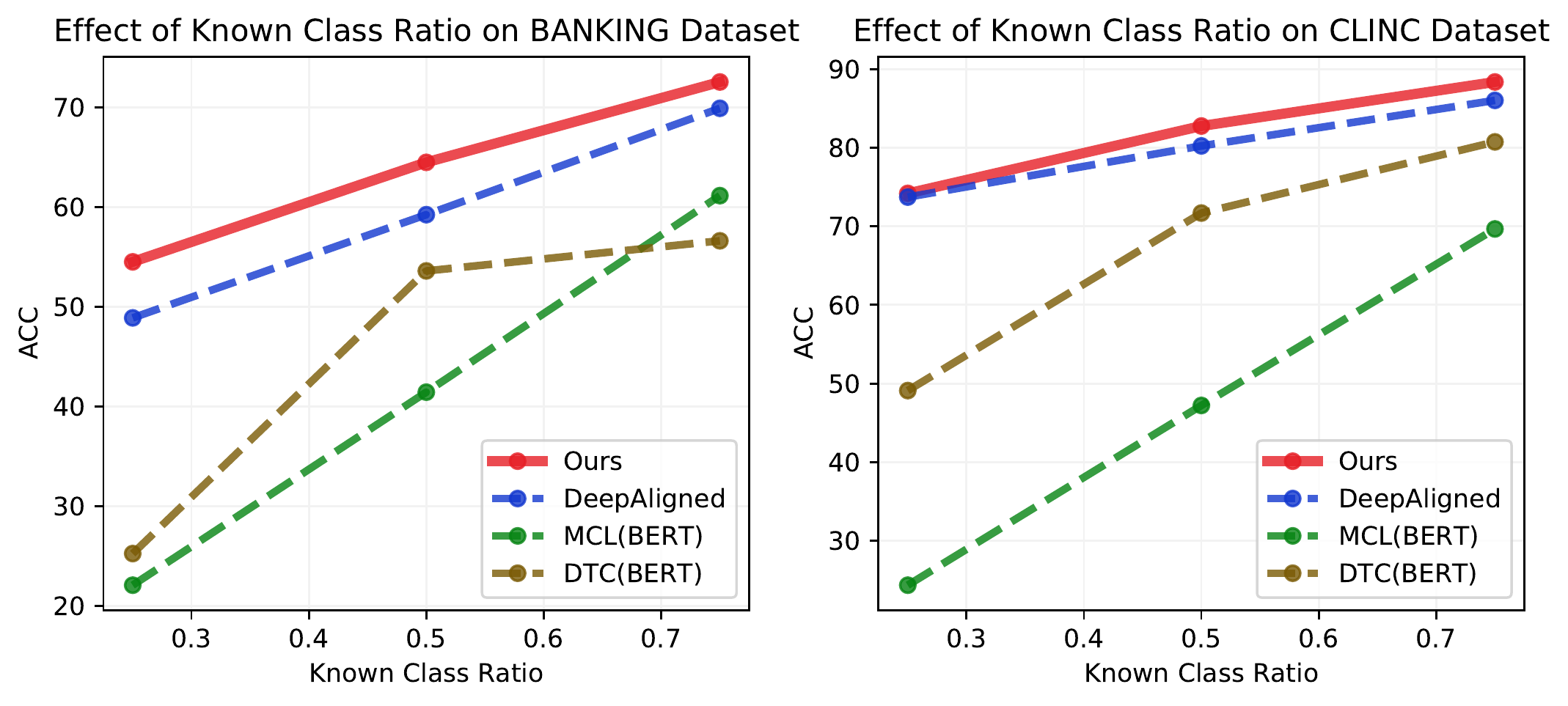}
    \caption{The effect of Known Class Ratio on datasets (Left: BANKING, Right: CLINC). The compared results are retrieved from~\protect\citet{zhang2021discovering}.}
    \label{fig:known class ratio}
\end{figure}

\subsection{Effect of the Known Intent Ratio}
We also investigate the effect of known intent ratios on the model. We adopt different known class ratios (25\%, 50\% and 75\%) and experiment on datasets. As shown in Figure~\ref{fig:known class ratio}, our method also shows better performance compared with other methods. Interestingly, The advantage of our method in dataset BANKING is particularly obvious. We speculate that this may be related to the imbalance of samples in each intent in the BANKING dataset. Although there are more known intents, it is unable to provide enough labeled and balanced samples. As a result, the previous methods (e.g. DeepAligned) not only failed to transfer more prior knowledge but also exacerbated the speed of forgetting in the follow-up process.

\begin{figure}[htb]
    \centering
    \includegraphics[width=\linewidth]{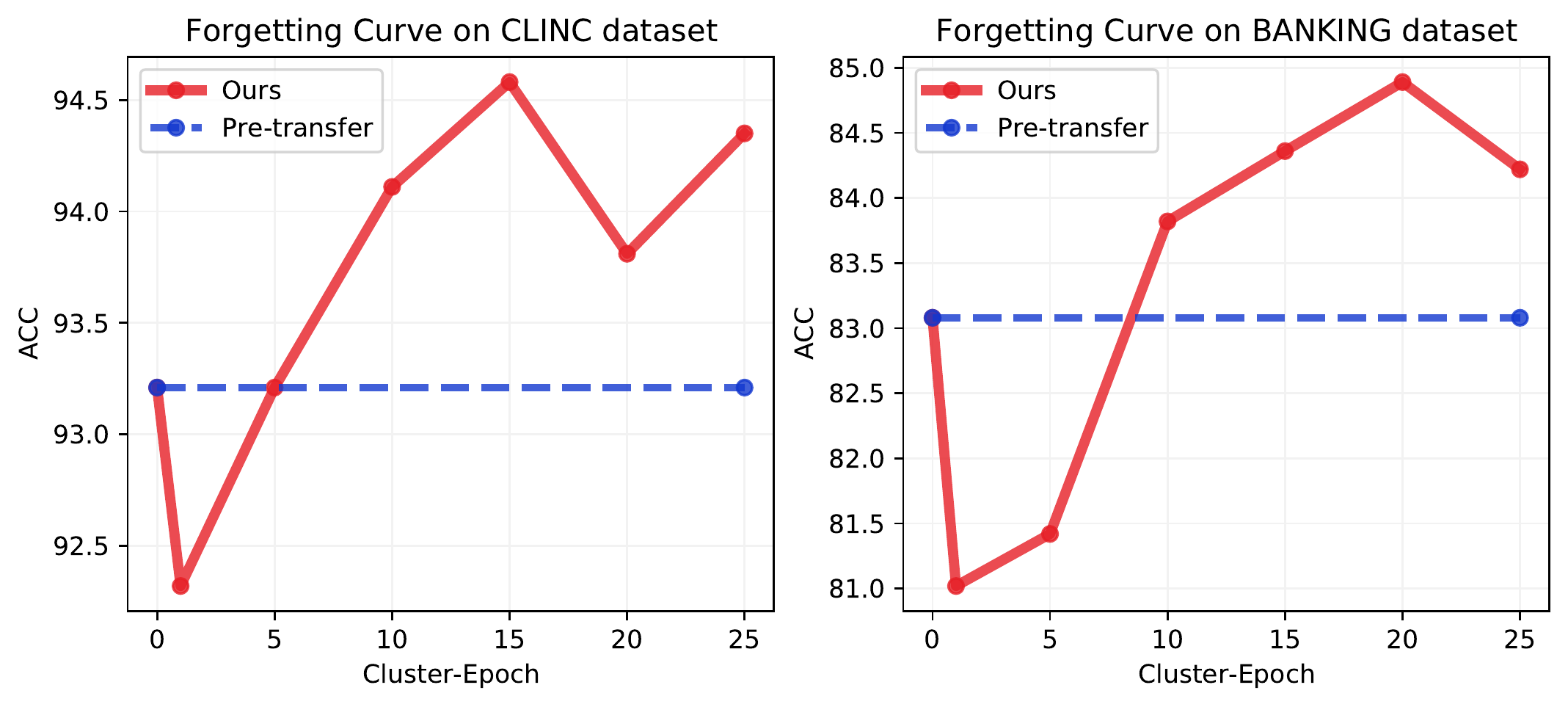}
    \caption{The knowledge curves of our method (Red). 
    During intent assignments in our method, we also test the performance of the model on the validation set used in the pre-transfer stage and show that with the advancement of clustering, our performance can be better than in the pre-transfer stage. The blue line represents the baseline obtained by the model in the pre-transfer stage.}
    \label{fig:our-probknowledge}
\end{figure}

\subsection{More Than Remembering Prior Knowledge}
\label{sec:prior Knowledge}
We showed knowledge forgetting in DeepAligned in introduction. Interestingly, there is a phenomenon called \textit{model shift} in the field of computer vision, which means fine-tuned model would shift towards labeled data, see~\citet{wang2021self} for details. However, in our scenario and our experiments, the model is moving in the opposite direction and is shifting away from labeled data, which reasonable explanation should be the forgetting of knowledge. After fine-tuning with labeled data, the prior knowledge is stored in the model in the form of model parameters. With the subsequent clustering steps, the parameters change gradually (the forgetting process is step by step from the forgetting curve). 

In our proposed framework, we just use a simple yet effective method (which can be dynamically adjusted according to different needs) to re-memorize this prior knowledge.
Therefore, as shown in Figure~\ref{fig:our-probknowledge}, we observe that our method does not have the catastrophic forgetting that occurs in DeepAligned. On the contrary, with the iteration (EM algorithm), our performance is better than that in the pre-transfer stage. We surmise that this is presumably because the knowledge contained in the unlabeled data corpus helps the identification of the known intents.

\section{Conclusion}
In this paper, we provide a probabilistic framework for intent discovery. This is the first complete theoretical framework for intent discovery. We also provide an efficient implementation based on this proposed framework. Compared with the existing methods, our method effectively alleviates the forgetting of prior knowledge transferred from known intents and provides intensive clustering supervised signals for discovering intents. Extensive experiments conducted in three challenging datasets demonstrate our method can achieve substantial improvements. The subsequent analysis also show that our method can better estimate the number of intents and adapt to various conditions. In the future, we will try different methods to perform intent assignments and explore more methods to approximate $p(Y^l|\mathcal{Z}, D^l;\theta)$ and $p(\mathcal{Z}|D^l\theta)$.


\bibliography{aaai23}

\begin{thebibliography}{25}
\providecommand{\natexlab}[1]{#1}

\bibitem[{Basu, Banerjee, and Mooney(2004)}]{basu2004active}
Basu, S.; Banerjee, A.; and Mooney, R.~J. 2004.
\newblock Active semi-supervision for pairwise constrained clustering.
\newblock In \emph{Proceedings of the 2004 SIAM international conference on
  data mining}, 333--344. SIAM.

\bibitem[{Caron et~al.(2018)Caron, Bojanowski, Joulin, and
  Douze}]{caron2018deep}
Caron, M.; Bojanowski, P.; Joulin, A.; and Douze, M. 2018.
\newblock Deep clustering for unsupervised learning of visual features.
\newblock In \emph{Proceedings of the European Conference on Computer Vision
  (ECCV)}, 132--149.

\bibitem[{Casanueva et~al.(2020)Casanueva, Tem{\v{c}}inas, Gerz, Henderson, and
  Vuli{\'c}}]{casanueva2020efficient}
Casanueva, I.; Tem{\v{c}}inas, T.; Gerz, D.; Henderson, M.; and Vuli{\'c}, I.
  2020.
\newblock Efficient intent detection with dual sentence encoders.
\newblock \emph{arXiv preprint arXiv:2003.04807}.

\bibitem[{Chang et~al.(2017)Chang, Wang, Meng, Xiang, and Pan}]{chang2017deep}
Chang, J.; Wang, L.; Meng, G.; Xiang, S.; and Pan, C. 2017.
\newblock Deep adaptive image clustering.
\newblock In \emph{Proceedings of the IEEE international conference on computer
  vision}, 5879--5887.

\bibitem[{Devlin et~al.(2018)Devlin, Chang, Lee, and
  Toutanova}]{devlin2018bert}
Devlin, J.; Chang, M.-W.; Lee, K.; and Toutanova, K. 2018.
\newblock Bert: Pre-training of deep bidirectional transformers for language
  understanding.
\newblock \emph{arXiv preprint arXiv:1810.04805}.

\bibitem[{Gowda and Krishna(1978)}]{gowda1978agglomerative}
Gowda, K.~C.; and Krishna, G. 1978.
\newblock Agglomerative clustering using the concept of mutual nearest
  neighbourhood.
\newblock \emph{Pattern recognition}, 10(2): 105--112.

\bibitem[{Hakkani-T{\"u}r et~al.(2013)Hakkani-T{\"u}r, Celikyilmaz, Heck, and
  Tur}]{hakkani2013weakly}
Hakkani-T{\"u}r, D.; Celikyilmaz, A.; Heck, L.; and Tur, G. 2013.
\newblock A weakly-supervised approach for discovering new user intents from
  search query logs.
\newblock In \emph{Annual Conference of the International Speech Communication
  Association (Interspeech)}.

\bibitem[{Hakkani-T{\"u}r et~al.(2015)Hakkani-T{\"u}r, Ju, Zweig, and
  Tur}]{hakkani2015clustering}
Hakkani-T{\"u}r, D.; Ju, Y.-C.; Zweig, G.; and Tur, G. 2015.
\newblock Clustering novel intents in a conversational interaction system with
  semantic parsing.
\newblock In \emph{Sixteenth Annual Conference of the International Speech
  Communication Association}.

\bibitem[{Han, Vedaldi, and Zisserman(2019)}]{han2019learning}
Han, K.; Vedaldi, A.; and Zisserman, A. 2019.
\newblock Learning to discover novel visual categories via deep transfer
  clustering.
\newblock In \emph{Proceedings of the IEEE/CVF International Conference on
  Computer Vision}, 8401--8409.

\bibitem[{Hsu, Lv, and Kira(2017)}]{hsu2017learning}
Hsu, Y.-C.; Lv, Z.; and Kira, Z. 2017.
\newblock Learning to cluster in order to transfer across domains and tasks.
\newblock \emph{arXiv preprint arXiv:1711.10125}.

\bibitem[{Hsu et~al.(2019)Hsu, Lv, Schlosser, Odom, and Kira}]{hsu2019multi}
Hsu, Y.-C.; Lv, Z.; Schlosser, J.; Odom, P.; and Kira, Z. 2019.
\newblock Multi-class classification without multi-class labels.
\newblock \emph{arXiv preprint arXiv:1901.00544}.

\bibitem[{Larson et~al.(2019)Larson, Mahendran, Peper, Clarke, Lee, Hill,
  Kummerfeld, Leach, Laurenzano, Tang et~al.}]{larson2019evaluation}
Larson, S.; Mahendran, A.; Peper, J.~J.; Clarke, C.; Lee, A.; Hill, P.;
  Kummerfeld, J.~K.; Leach, K.; Laurenzano, M.~A.; Tang, L.; et~al. 2019.
\newblock An evaluation dataset for intent classification and out-of-scope
  prediction.
\newblock \emph{arXiv preprint arXiv:1909.02027}.

\bibitem[{Lin, Xu, and Zhang(2020)}]{lin2020discovering}
Lin, T.-E.; Xu, H.; and Zhang, H. 2020.
\newblock Discovering new intents via constrained deep adaptive clustering with
  cluster refinement.
\newblock In \emph{Proceedings of the AAAI Conference on Artificial
  Intelligence}, volume~34, 8360--8367.

\bibitem[{MacQueen et~al.(1967)}]{macqueen1967some}
MacQueen, J.; et~al. 1967.
\newblock Some methods for classification and analysis of multivariate
  observations.
\newblock In \emph{Proceedings of the fifth Berkeley symposium on mathematical
  statistics and probability}, volume~1, 281--297. Oakland, CA, USA.

\bibitem[{Padmasundari(2018)}]{padmasundari2018intent}
Padmasundari, S.~B. 2018.
\newblock INTENT DISCOVERY THROUGH UNSUPERVISED SEMANTIC TEXT CLUSTERING.
\newblock \emph{Proc. Interspeech 2018}, 606--610.

\bibitem[{Reimers and Gurevych(2019)}]{reimers2019sentence}
Reimers, N.; and Gurevych, I. 2019.
\newblock Sentence-bert: Sentence embeddings using siamese bert-networks.
\newblock \emph{arXiv preprint arXiv:1908.10084}.

\bibitem[{Shen et~al.(2021)Shen, Sun, Zhang, and Najmabadi}]{shen2021semi}
Shen, X.; Sun, Y.; Zhang, Y.; and Najmabadi, M. 2021.
\newblock Semi-supervised Intent Discovery with Contrastive Learning.
\newblock In \emph{Proceedings of the 3rd Workshop on Natural Language
  Processing for Conversational AI}, 120--129.

\bibitem[{Shi et~al.(2018)Shi, Chen, Sha, Li, Sun, Wang, and
  Zhang}]{shi-etal-2018-auto}
Shi, C.; Chen, Q.; Sha, L.; Li, S.; Sun, X.; Wang, H.; and Zhang, L. 2018.
\newblock Auto-Dialabel: Labeling Dialogue Data with Unsupervised Learning.
\newblock In \emph{Proceedings of the 2018 Conference on Empirical Methods in
  Natural Language Processing}, 684--689. Brussels, Belgium: Association for
  Computational Linguistics.

\bibitem[{Wang et~al.(2021)Wang, Gao, Long, and Wang}]{wang2021self}
Wang, X.; Gao, J.; Long, M.; and Wang, J. 2021.
\newblock Self-tuning for data-efficient deep learning.
\newblock In \emph{International Conference on Machine Learning}, 10738--10748.
  PMLR.

\bibitem[{Xie, Girshick, and Farhadi(2016)}]{xie2016unsupervised}
Xie, J.; Girshick, R.; and Farhadi, A. 2016.
\newblock Unsupervised deep embedding for clustering analysis.
\newblock In \emph{International conference on machine learning}, 478--487.
  PMLR.

\bibitem[{Xu et~al.(2015)Xu, Wang, Tian, Xu, Zhao, Wang, and Hao}]{xu2015short}
Xu, J.; Wang, P.; Tian, G.; Xu, B.; Zhao, J.; Wang, F.; and Hao, H. 2015.
\newblock Short text clustering via convolutional neural networks.
\newblock In \emph{Proceedings of the 1st Workshop on Vector Space Modeling for
  Natural Language Processing}, 62--69.

\bibitem[{Yang et~al.(2017)Yang, Fu, Sidiropoulos, and Hong}]{yang2017towards}
Yang, B.; Fu, X.; Sidiropoulos, N.~D.; and Hong, M. 2017.
\newblock Towards k-means-friendly spaces: Simultaneous deep learning and
  clustering.
\newblock In \emph{international conference on machine learning}, 3861--3870.
  PMLR.

\bibitem[{Zhang et~al.(2021)Zhang, Xu, Lin, and Lyu}]{zhang2021discovering}
Zhang, H.; Xu, H.; Lin, T.-E.; and Lyu, R. 2021.
\newblock Discovering new intents with deep aligned clustering.
\newblock In \emph{Proceedings of the AAAI Conference on Artificial
  Intelligence}, volume~35, 14365--14373.

\bibitem[{Zhang et~al.(2022)Zhang, Zhang, Zhan, Wu, and
  Lam}]{zhang-etal-2022-new}
Zhang, Y.; Zhang, H.; Zhan, L.-M.; Wu, X.-M.; and Lam, A. 2022.
\newblock New Intent Discovery with Pre-training and Contrastive Learning.
\newblock In \emph{Proceedings of the 60th Annual Meeting of the Association
  for Computational Linguistics (Volume 1: Long Papers)}, 256--269. Dublin,
  Ireland: Association for Computational Linguistics.

\bibitem[{Zhou, Liu, and Qiu(2022)}]{zhou-etal-2022-knn}
Zhou, Y.; Liu, P.; and Qiu, X. 2022.
\newblock {KNN}-Contrastive Learning for Out-of-Domain Intent Classification.
\newblock In \emph{Proceedings of the 60th Annual Meeting of the Association
  for Computational Linguistics (Volume 1: Long Papers)}, 5129--5141. Dublin,
  Ireland: Association for Computational Linguistics.

\end{thebibliography}

\end{document}